# An Autoencoder-based Snow Drought Index

**Sinan Rasiya Koya¹, Kanak Kanti Kar¹, Shivendra Srivastava¹, Tsegaye Tadesse², Mark Svoboda², Tirthankar Roy¹,***

¹Department of Civil and Environmental Engineering, University of Nebraska-Lincoln
²National Drought Mitigation Center, University of Nebraska-Lincoln

***Corresponding Author:** Tirthankar Roy (roy@unl.edu)

## ABSTRACT

In several regions across the globe, snow has a significant impact on hydrology. The amounts of water that infiltrate the ground and flow as runoff are driven by the melting of snow. Therefore, it is crucial to study the magnitude and effect of snowmelt. Snow droughts, resulting from reduced snow storage, can drastically impact the water supplies in basins where snow predominates, such as in the western United States. Hence, it is important to detect the time and severity of snow droughts efficiently. We propose Snow Drought Response Index or SnoDRI, a novel indicator that could be used to identify and quantify snow drought occurrences. Our index is calculated using cutting-edge ML algorithms from various snow-related variables. The self-supervised learning of an autoencoder is combined with mutual information in the model. In this study, we use random forests for feature extraction for SnoDRI and assess the importance of each variable. We use reanalysis data (NLDAS-2) from 1981 to 2021 for the Pacific United States to study the efficacy of the new snow drought index. We evaluate the index by confirming the coincidence of its interpretation and the actual snow drought incidents.

## 1. Introduction

In many regions worldwide, snow has a vital contribution to drought occurrence [1], evident from the role of snow in regional and global water resources and climate [2-4]. Recently, it has led to a broad discussion on the association between droughts and snow, along with the emergence of several studies focusing on "snow drought," indicative of lower-than-normal snow conditions [1,5-9]. However, a consistent way of characterizing snow droughts is missing in these past studies, resulting in the absence of a solid framework to detect snow droughts.

Different authors defined snow droughts differently, making the analysis of these droughts conditioned upon and potentially sensitive to the definitions. For example, a recent study assessing the global snow drought hotspots and characteristics considered a snow drought event as a deficit of snow water equivalent (SWE) [1]. Another study argued that defining snow droughts just in terms of SWE might not be sufficient, and it referred to snow drought as a combination of general droughts and shortages in snow storage, reflecting both the lack of winter precipitation and SWE [5]. Subsequently, Hatchett & McEvoy (2018) defined different types of snow droughts based on the origination, persistence, and termination of below-normal snow accumulations [9]. Later, several studies expressed snow droughts in terms of threshold percentiles, which is essentially subjective [8,10]. Some of these studies used average SWE to identify snow droughts, whereas others used maximum SWE [8,10]. Although we need an index to study and predict snow droughts, the lack of coherence in the characterization of snow droughts potentially questions the reliability of a snow drought index based on strict definitions. Therefore, we need a framework to calculate a snow drought index independent of such definitions, which, at the same time, can also capture the signals of a snow drought.

To date, only a limited number of studies have been conducted on snow drought indices. The recently developed Standardized Snow Water Equivalent Index (SWEI) is obtained through the inverse standard normal distribution of probabilities associated with SWE integrated over three months [1]. Keyantash & Dracup (2004) developed the aggregated drought index (ADI) based on rainfall, evapotranspiration, streamflow, reservoir storage, soil moisture, and snow water content variable [11]. Here, a principal component analysis (PCA) was employed to reduce the dimensionality and explain the key variabilities represented by the selected variables. Staudinger et al. (2014) developed an extension to the Standardized Precipitation Index (SPI) named Standardized Snow Melt and Rain Index (SMRI), where they used the sum of rainfall and snowmelt variables instead of precipitation [12]. Qiu (2013) modified the standard Palmer Drought Severity Index (PDSI; Palmer, 1965) by including degree day factor (DDF), an empirical threshold temperature-based snowmelt model, to account for snow processes [13,14]. This modification improved the drought monitoring capabilities in several snow-dominated regions [13,15]. However, these indices often depend upon in-situ observations, which might not be readily available in many regions, and the problem is exacerbated in ungauged or sparsely gauged regions. This calls for an index that can leverage remote sensing datasets and bypass the need for extensive in-situ observation networks.

Merging and extracting necessary information on snow droughts from a wide range of variables present in remote sensing datasets can be challenging since not all variables are equally related to the formation of snow droughts. However, we need to identify important variables so that they can be merged to form one single index. Machine learning (ML)-based feature selection algorithms are promising in this regard since they can filter our variables based on their importance [16-19]. Thus, we can use ML techniques to infer the influence of hydroclimatic variables on snow droughts. Apart from this, information theory-based methods can manifest the relative influence of variables and their causal connections [20-22]. Mutual Information (MI), a measure of how much information about one random variable is contained in another random variable, has been widely applied in feature selection problems [23,24].

In this study, we are introducing a new snow drought index, Snow Drought Response Index (SnoDRI), using a combination of ML techniques and MI. SnoDRI considers several snow factors while assessing snow drought, including SWE and snow fraction. Importantly, to estimate SnoDRI, we do not require any ground measurements. Our results show that SnoDRI could successfully detect the signals of a historical snow drought event. Moreover, our framework could provide insights into the crucial features impacting the occurrence of snow drought.

## 2. Methods

### 2.1. Study Area

The Western United States is characterized as a snow drought hot spot where snow drought becomes widespread, intensified, and prolonged [1]. Snow plays a significant role in the hydrology of the Western United States. Recent events have shown that the water resources and management in this region are hugely influenced by snow droughts [5]. Therefore, this study focused on three states (Pacific Coast States) in the western USA: Washington, California, and Oregon (Figure 1). The highest elevation of this region ranged from 1,547 m in California to 520 m in Washington State. As a result, elevation and climate variables associated with orographic precipitation and the variability in average annual maximum snow water accumulation are considerable.

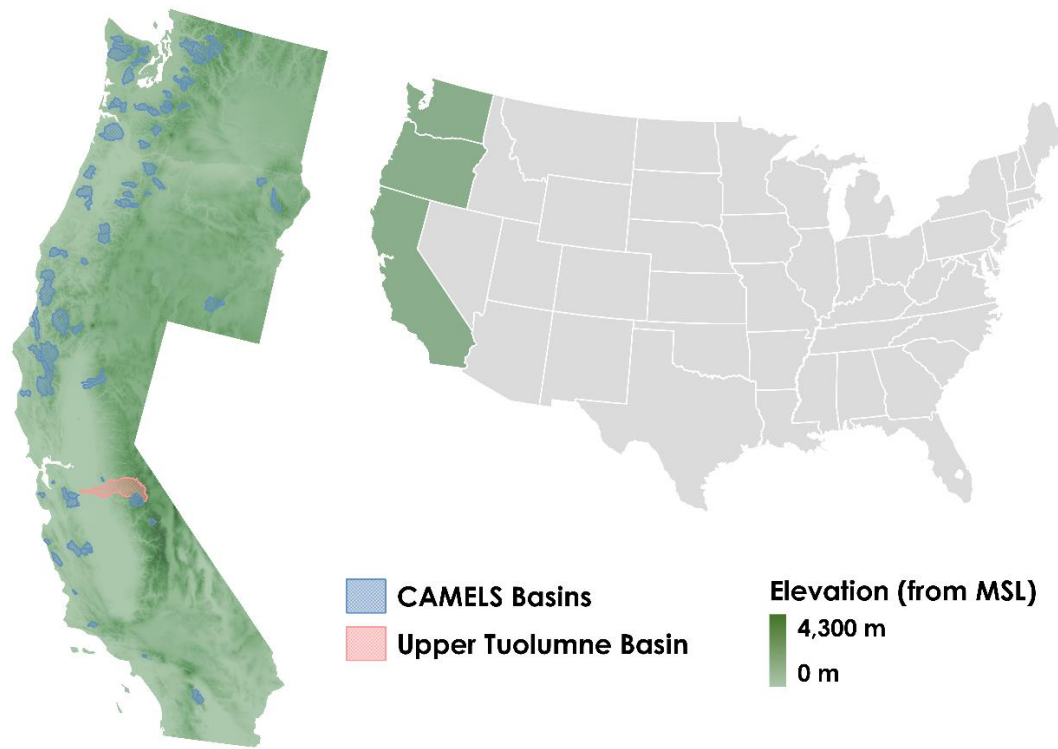

*Figure 1.* The Pacific Coast States of the United States showing the study basins.

To validate our results, we selected a mountainous area situated in the upper Tuolumne River basin of the Sierra Nevada [25]. About 60% of water in Southern California is received from the Sierra Nevada snowpack [26]. As such, the snowpack plays a vital role in this region, which was confirmed by the impacts of below-normal snow conditions on water resources, ecosystems and recreation [8]. Due to the mountainous region and varied elevation, individual climatological extreme events can produce a drastically different response in magnitude and spatial variability of the snowpack in this region [27]. Recently, 2014 and 2015 showed a lack of snow accumulation and winter precipitation, indicating snow drought events in the upper Tuolumne basin [5,8,9]. The historical precipitation trends in this region have shifted from snow to rain, resulting in more frequent droughts [10,28]. This shift has impacted headwater hydrology and downstream reservoir management of basins in the Sierra Nevada [29].

## 2.2. Data used for SnoDRI

### NLDAS-2 Variables

The North American Land Data Assimilation System (NLDAS) is a multi-institutional partnership project intended to develop land-surface model datasets from observations and reanalysis with quality control that is coherent across space and time [30]. NLDAS data is comprised of hourly data in gridded format with a spatial resolution of 0.125 ° x 0.125 °. An improved version, NLDAS-2, was later developed by determining and rectifying existing errors in forcing data and models [31]. NLDAS-2 changed data sources and their inherent biases, upgraded the model along with recalibrated parameters, and increased the period of forcing data and simulations [31]. NLDAS-2 data provides a total of eleven variables, given in Table 1. All these variables are spatially aggregated for interested basins and converted to monthly timeseries.

**Table 1.** List of variables considered for analysis.

| Variables name | Resolution | Period | Sources |
|---|---|---|---|
| U wind component (m/s) at 10 meters above surface | 0.125° × 0.125° | 1979-2020 Hourly | NLDAS 2 |
| V wind component (m/s) at 10 meters above surface | | | |
| Air temperature (K) at 2 meters above surface | | | |
| Specific humidity (kg/kg) at 2 meters above surface | | | |
| Surface pressure (Pa) | | | |
| Surface downward longwave radiation (W/m²) | | | |
| Surface downward shortwave radiation (W/m²) | | | |
| Precipitation hourly total (kg/m²) | | | |
| Fraction of total precipitation that is convective | | | |
| Convective available potential energy (J/kg) | | | |
| Potential evaporation (kg/m^2) | | | |
| Snow water equivalent | 4km × 4km | 1982-2020 Daily | NSIDC |
| Discharge | Point | 1979-2020 Daily | USGS |

## Standardized Precipitation Index (SPI)

The Standardized Precipitation Index (SPI) is an index widely used for quantifying precipitation anomalies (McKkee et al., 1993). The SPI is obtained by mapping the actual probability distribution of precipitation to a normal distribution. A zero SPI indicates normal conditions. The positive values of SPI represent wet conditions, whereas the negative values represent dry conditions. The larger the negative value of SPI, the higher the severity of drought conditions. We calculated SPIs at three timescales (3, 4, and 6 months) and provided them as inputs for the SnoDRI. The SPI at 3, 4, and 6 months timescales were chosen because the snow processes and the impact of reduced winter precipitation generally occur at these timescales. These SPIs can reflect the reduced snowpack and discharge from snowmelt. For the same reason, SPIs at longer timescales (such as SPI-12 and 60) are potentially unimportant for snow droughts. Therefore, we have also fed the model with SPI-12 and SPI-60 as a sanity check to see whether the model can discard irrelevant information, which is confirmed by our results (see section 4.2). The SPI calculation only requires precipitation; it smoothens the time series data and maps the actual distribution of precipitation to normal distribution. We used the basin aggregated NLDAS-2 precipitation for computing SPIs.

## Snow Water Equivalent

We use snow water equivalent (SWE) as an indicator variable for the snow drought case used for validation. SWE is the target variable for the random forest model (discussed in a later section) used in selecting the input features for our index. We obtained SWE data and snow depth from assimilated in-situ and modeled data over the conterminous US, Version 1 [33,34] from National Snow and Ice Data Center (NSIDC). This data provides daily SWE and snow depth at a spatial resolution of 4km x 4km for the conterminous United States (CONUS). We collected the SWE data from 1982 to 2020 and spatially aggregated them for the study basins and converted them to monthly timeseries.

## CAMELS Basin Shapefiles

Catchment Attributes and MEteorology for Large-sample Studies (CAMELS) provide the time series of meteorological forcings and attributes of 671 basins across the CONUS [35,36]. These basins are least affected by human actions [36]. The dataset contains different categories of basin attributes: topography, climate, streamflow, land cover, soil, and geology [36]. We used the basin shapefiles provided in the dataset and filtered basins (a total of 85) belonging to the Pacific Coast states. The gridded datasets are aggregated to the basin scale using these shapefiles.

## Discharge

The discharge data for these basins were collected from the US Geological Survey's (USGS) streamflow measurements provided in the CAMELS dataset. USGS collects, monitors, and analyzes existing resource conditions across the different sites in the US. USGS stations measure velocity through a current meter or acoustic doppler current profiler. The discharge is computed by multiplying the cross-sectional area by the estimated velocity. For this study, daily data were obtained from 1979 through 2020. The daily flow records for USGS gage stations provide the value of mean discharge for each day of the water year (US Geological Survey, 2023).

## 2.3. SnoDRI Framework

In this work, as shown in Figure 2, we followed a general framework where all input data were standardized up front. Using this dataset, we found the weight of each input variable from an ML model coupled with MI. The details of these components are discussed in the later on. Once the weights are obtained, each variable is multiplied with the corresponding weight, and the weighted inputs are added. Thus, the resultant values are standardized to obtain the SnoDRI index.

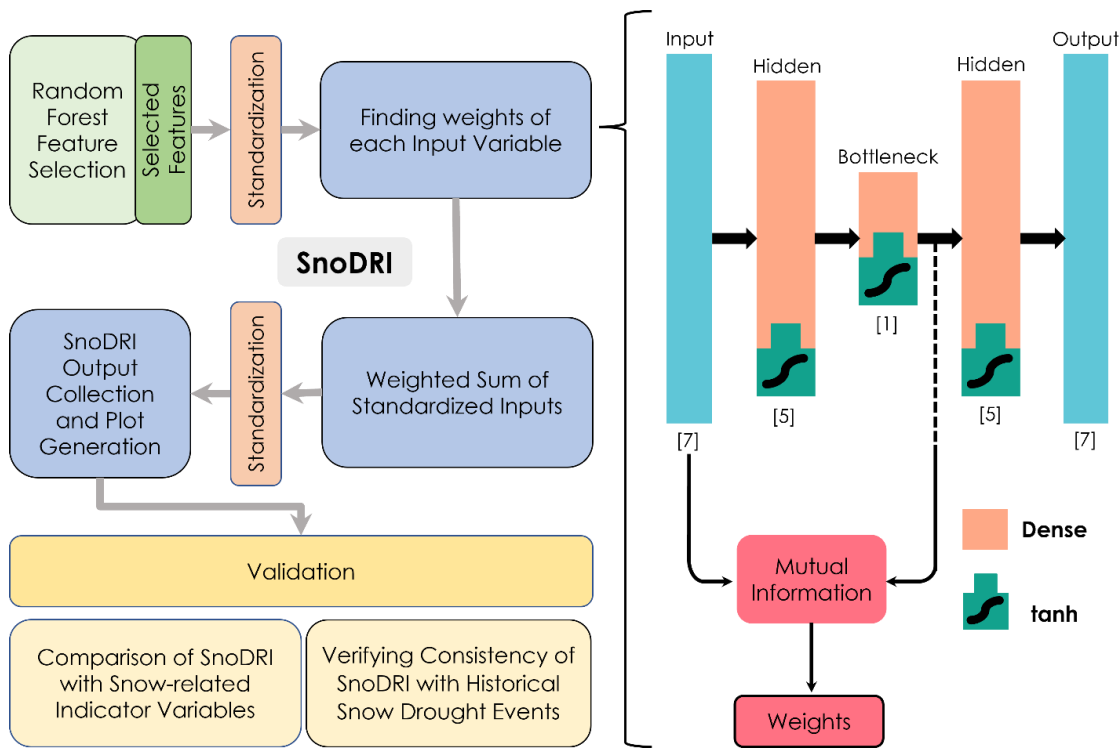

*Figure 2. The methodology used for developing SnoDRI. On the left-hand side is the flowchart with the steps followed in this study. On the right-hand side is the framework with autoencoder (top) and Mutual Information used to calculate the weight of each input variable. Inside the square bracket is the number of nodes in each layer of the autoencoder.*

## 2.4. Random Forest Regression

Random Forests are a collection of decision trees. Each tree in a Random Forest is built based on a random subset of variables from bootstrapped data, which can be used for classification and regression problems [38]. Since every regression tree inside the random forest identifies an order of variables to classify the training dataset, we can leverage the random forest regression algorithm to find the feature importance of training variables in predicting the target variable. In this study, we use random forest regression to select the important variables to develop SnoDRI. We aggregated the NLDAS-2 variables for 85 CAMELS basins in the Western US and regressed it against SWE and discharge within the basins. This yields two sets of feature importance corresponding to SWE and discharges for each basin. Then by taking the mean of all basins, the average feature importance of variables for the Western US is calculated. The union of the top four variables in average feature importance for predicting SWE and discharge is used to compute SnoDRI.

## 2.5. Self-Supervised Learning with Autoencoder

Given the fact that we are trying to develop a novel snow drought index derived from different snow variables, we do not have a target variable to train a model. The absence of a target variable makes our problem a case of unsupervised learning. We used the ML algorithm of autoencoders, a particular type of neural network that is used for dimensionality reduction [39]. During the learning process, autoencoders "discard" the insignificant information present in the dataset. An autoencoder consists of an input layer, an encoder with hidden layers, a bottleneck layer, a decoder with hidden layers, and an output layer that tries to reconstruct the input data (self-supervised learning). The NLDAS-2 variables identified through random forest regression (section 3.1) are initially passed to the input layer. As the data passes through the encoder and reaches the bottleneck layer, it encodes the entire data into a reduced form which can be regarded as a 'compressed' form of important information in the entire dataset. After several trial-and-error iterations, we finalized an autoencoder network with three hidden layers, including a bottleneck. The structure of the autoencoder network is shown in Figure 2. The bottleneck layer consists of one neuron, and the other two hidden layers consist of fifteen neurons. The hidden layers and the bottleneck layer used tanh activation functions which take care of the nonlinearities in the model. We found that the Adam optimization algorithm with the Huber loss function gives better training of our model. The training of the model in 3000 epochs as single batches provided the best possible accuracy with the given datasets. Loss and accuracy stayed nearly the same regardless of the additional increase in the number of epochs.

A valid question here is why we cannot directly use the weights of the trained autoencoder. To explain this, referring to Figure 3, we must look at all possible trajectories of "information flow" from one input variable to the compressed bottleneck output. We can see that other input variables also influence the weights. For example, the highlighted path in Figure 3 shows possible trajectories of "information flow" from input X1 before it reaches the compressed bottleneck output. Since the nodes in the second layer are connected to all input nodes from previous layers, weights $W_{11}^{(2)}$ and $W_{12}^{(2)}$ are optimized based on the information from all input variables. As the information from each input node is divided and passed to all nodes in the hidden layer, weights $W_{11}^{(1)}$ and $W_{12}^{(1)}$ are optimized based on the division of information from input nodes. Therefore, the weights inside the autoencoder network are not representative of the relative contribution of each input to the compressed bottleneck output. This issue led to using an alternative method, MI, to infer the weights for each input feature.

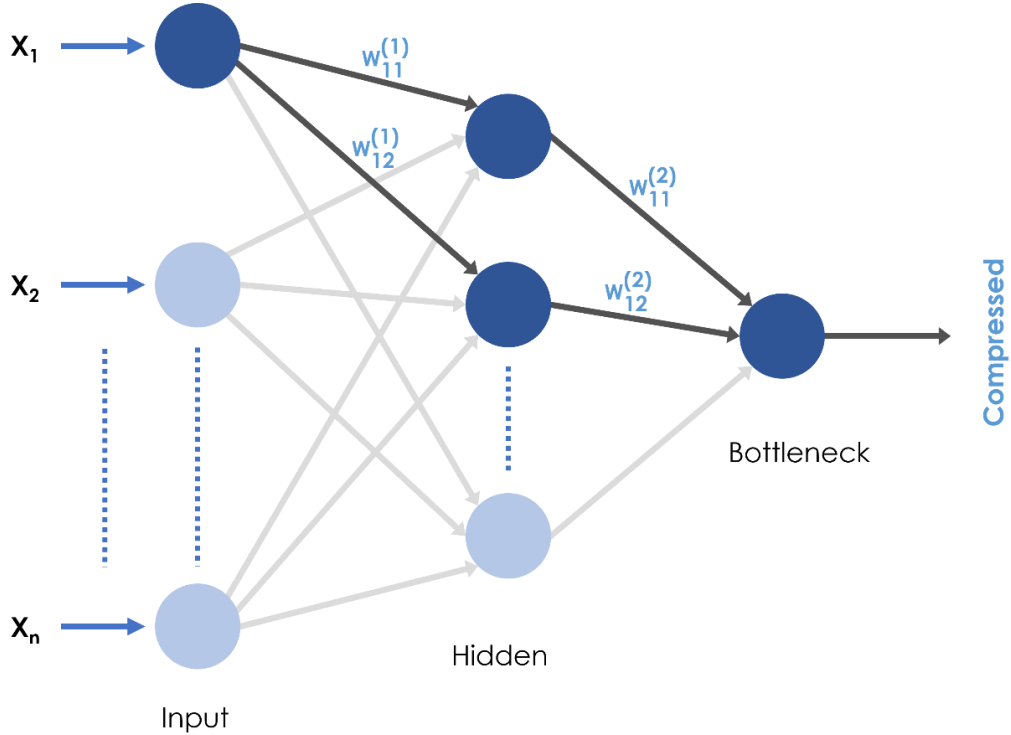

*Figure 3. The encoder portion (first half) of the autoencoder neural network. The highlighted paths show two possible "information flows" (black arrowed lines) from the first input feature to the compressed bottleneck output.*

## 2.6. Mutual Information

Mutual Information (MI) is a measure of how much information, on average, can one random variable tell us about another random variable [22,40,41]. It can be conceptualized as the reduction in entropy of one variable, given the information about another variable [21]. MI between two random variables, X and Y, expressed as I(X;Y), is calculated using Equation 1 [22,40]. Here $P_X(x)$ and $P_Y(y)$ are marginal probabilities, and $P_{XY}(x,y)$ is the joint probability.

$$I(X;Y) = \sum P_{XY}(x,y) \log \frac{P_{XY}(x,y)}{P_X(x)P_Y(y)} = E_{P_{XY}} \log \frac{P_{XY}}{P_X P_Y} \qquad (1)$$

This study used MI between the bottleneck output and each input variable to determine their weights. As shown in Figure 2, the weights of each variable are multiplied by themselves and added (weighted sum), which is then standardized to obtain SnoDRI values. Since the bottleneck represents a "compressed" version of all input data, the MI shows how much of each variable is contained inside the "compressed" form.

## 2.7. Rain-Snow Partitioning

We also used snow fraction as an input into the model. In the modeling realm, there are different rain-snow partitioning schemes based on which precipitation variable is used as the forcing is separated into snow and rainfall. Classifying rain and snow based on a threshold temperature is the most straightforward scheme. This method is susceptible to the choice of threshold temperature. In another scheme proposed by Jordan (1991), the snow percent is calculated as a linear stepwise function of air temperature[42]. In this study, we estimated the snow fraction based on a sigmoid function of wet bulb temperature, as Wang et al. (2019) proposed [43]. The wet bulb temperature is calculated from air temperature, specific humidity, and surface pressure from the NLDAS-2 dataset.

## 2.8. Validation

There is no single framework for validating a drought index. Based on the purpose of the new index, one must examine whether the index follows the drought indicators, such as the scarcity of relevant environmental variables. This study uses the anomaly in SWE as an indicator variable. A negative anomaly represents the lack of snow accumulation compared to normal and vice versa. The lower the anomaly, the more severe the snow drought. The discharge co-occurring with SWE is another indicator variable. Reduced discharge due to low meltwater contribution can be a potential impact of the snow drought. A lower discharge following a lower SWE is a prime case of snow drought.

We compared the novel SnoDRI with patterns of SWE and discharge in the Upper Tuolumne basin of California. Hatchett et al. (2022) and Harpold & Dettinger (2017) reported a snow drought in this region during the winter (Jan to Apr) of 2014 and 2015[5,8]. A lower SWE, along with a lower discharge, is taken as a signal of snow drought in the basin. We checked if these signals correspond to a lower value of SnoDRI. Meanwhile, the index should not give a false positive forecast.

## 3. Results

## 3.1. Feature Importance

Through random forest regressions for 85 CAMELS basins, we obtained the average feature importance of NLDAS-2 variables for the Pacific Coast States of the US in predicting SWE and discharge. This method gives a sense of the relative significance of each variable in generating SWE or discharge. Figure 4a shows the feature importance for predicting SWE. We see that temperature is the top significant variable in determining SWE. This is most likely because the temperature decides the amount of snowfall (vs. rainfall) in the precipitation. Several rain-snow partitioning schemes are highly sensitive to temperature [42-45]. After temperature, the downward shortwave radiation affects the SWE most. Since the primary source of energy that drives the atmospheric process is the incoming shortwave radiation, we can expect that the formation and accumulation of snow are highly dependent on shortwave radiation. Specific humidity is the third most crucial variable for SWE as obtained from random forest regression. Specific humidity,

(the measure of water content in the atmosphere) could have a significant effect on the formation of snow.

Figure 4b indicates the average feature importance for predicting discharge for the Pacific Coast States of the US. The results show that precipitation has a very high significance for estimating discharge in the basin. It is intuitive that the incoming precipitation, in the form of rainfall or snow, would contribute the most to generating river runoff as precipitation is the primary water source for any basin. Though temperature, zonal wind, and downward shortwave radiation are most important after precipitation, their importance is far lower than precipitation, as the results of random forest regression suggest.

From both cases, random forest regression targeting SWE and discharge, we identified the top three variables having the highest average feature importance for our study area. The union of these variables gives a set of APCP, TMP, DSWRF, SPFH, and VGRD, which are used as input variables in SnoDRI calculations.

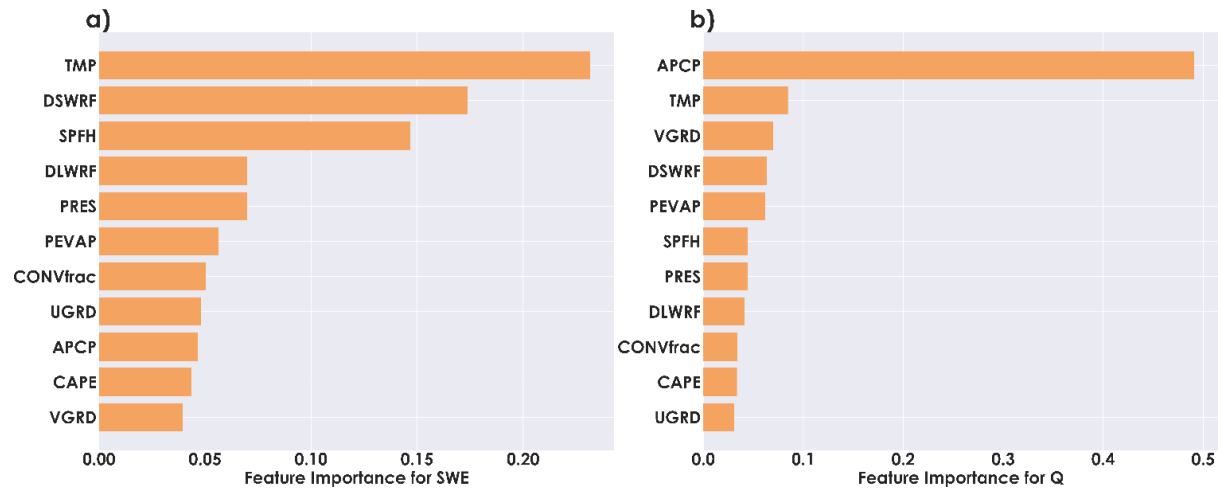

*Figure 4.* Average Feature importance for SWE (a) and Q (b) in basins of pacific states. The long names of variables are provided in Table 1.

### 3.2. Weights from Mutual Information

The MI between the compressed bottleneck output and each input variable measured the relative importance of the corresponding variable in the compressed data (bottleneck output). Figure 5 shows the approach for obtaining weights and the subsequent results. We can see that the downward shortwave radiation, SPIs, temperature, and snow fraction are found to be carrying more weight than other variables. Downward shortwave radiation from the sun entering the atmosphere, besides acting as the sole energy source of hydroclimatic processes, causes the direct melting of snow. This shows how important the downward shortwave radiation can be in the occurrence of snow droughts, which is reflected as the highest weight estimated by our framework. However, many of the previous studies on snow droughts have not considered the impact of downward shortwave radiation in their assessment [1,7,12]. Our result suggests further investigating the direct and indirect association between downward shortwave radiation and snow droughts. The model also identified shorter-scale SPIs (3, 4, and 6 months) as significant. Since SPIs represent the lack of precipitation in the case of drought, they can be a proxy for lower snowfall and snowpack conditions. Following this, the model assigned more weight to temperature, complying with the fact that temperature drives snow processes from the formation of snow to the melting of snow. Snow fraction, another variable with considerable weight, can be directly related to the amount of snowpack. The lower the snow fraction, the lower the snowfall, leading to snow drought conditions. Interestingly, the model identifies that precipitation as such does not have a severe influence though the snow fraction partially contains precipitation information. The reason could be that the SPIs, which are the precipitation mapped to a normal distribution, already contain the information of precipitation. This can also be attributed to the ability of the model to dismiss redundant information. The lowermost weights were assigned to the SPI-12 and 60, two variables presupposed to be irrelevant for snow droughts, additionally confirmed the aforementioned ability of the model.

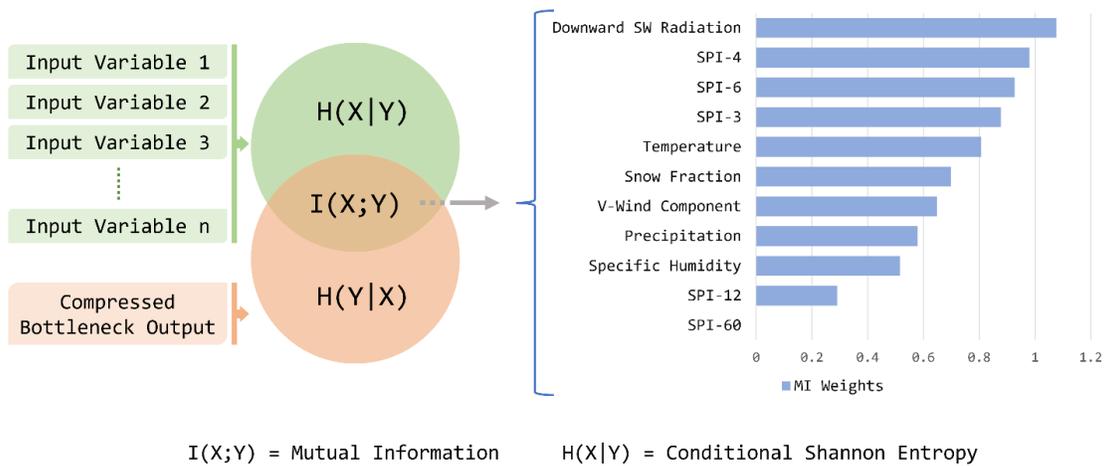

*Figure 5. Weights obtained for each input variable as the Mutual Information in each variable and the compressed bottleneck output.*

### 3.3. SnoDRI Evaluation

With the newly developed snow drought index, SnoDRI, we analyzed the conditions of snow drought in the Upper Tuolumne River basin from February 2013 to May 2019. This period is included in the evaluation period. In other words, the model has never seen the input dataset of this duration throughout its training. Over this evaluation period, the new index shows a good performance in capturing snow drought events. Figure 6 shows the SnoDRI calculated for Upper Tuolumne Basin in California, indicating a lower value corresponding to the reported snow droughts during the winter of 2013/14 and 2014/2015. Negative (positive) values of the SnoDRI suggest the presence (absence) and severity of the snow drought.

In addition, we placed SnoDRI and the potential signals of the snow drought in juxtaposition. Comparing SnoDRI with the anomaly in SWE, SnoDRI shows lower (more negative) values when the SWE anomaly is shallow. For instance, in 2014, 2015 and 2018, the Upper Tuolumne Basin showed a negative anomaly in SWE (Figure 6). During these years, SnoDRI showed lower values, indicating the occurrence of snow drought. On the other hand, during 2017, the region observed higher snow accumulation, leading to a positive SWE anomaly. SnoDRI indicates a continuous positive value during this period. Moreover, the higher (in 2017) and lower (in 2014, 2015, and 2018) values of original SWE and snow depth are reflected in SnoDRI. A lower discharge accompanied by lower SWE is another indicator of snow drought. In Figure 6, during all winters from 2013 to 2019, except for 2017 and 2019, the Upper Tuolumne basin produced low discharges. SnoDRI identifies this signal. Whereas in 2017 and 2019, the basin received a higher snow accumulation, and as a result the basin generated a higher discharge. This absence of snow drought was reflected in the SnoDRI, as seen in Figure 6. The higher the temperature, the higher the chance of snow droughts, leading to reduced snowfall and rapid snowmelt. In accordance with Figure 6, negative SnoDRI matched with positive anomalies in temperature and vice versa.

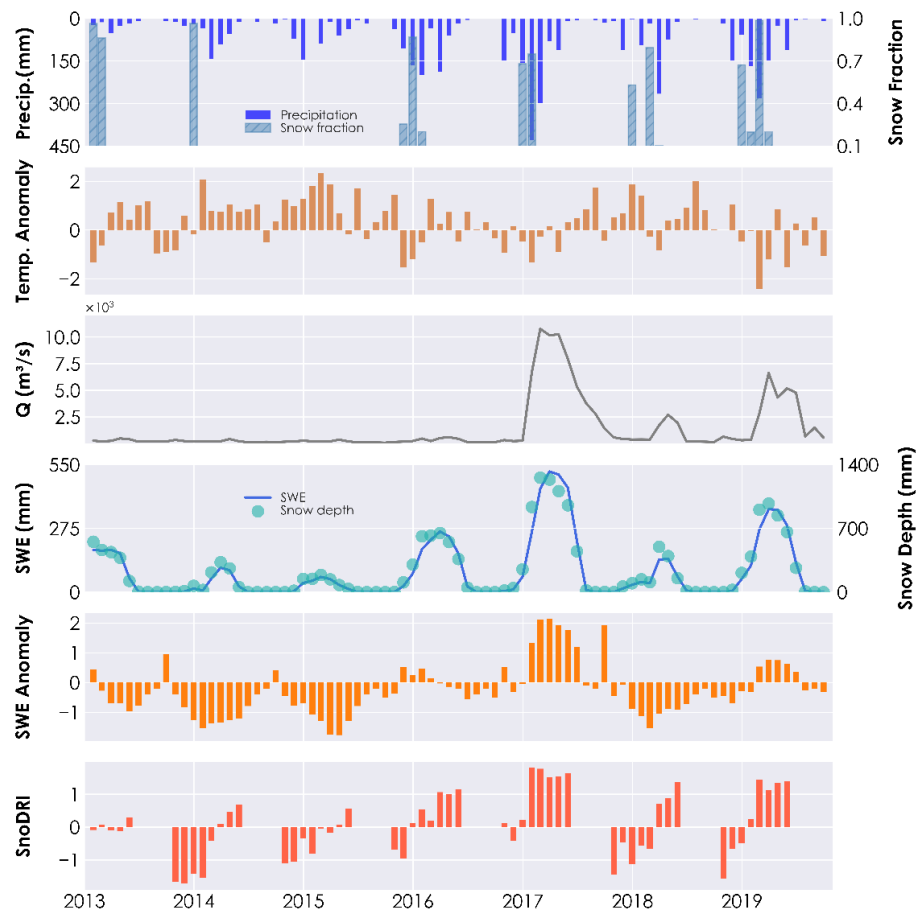

*Figure 6.* SnoDRI, SWE anomaly, SWE, and discharge in the Upper Tuolumne basin during the validation period.

## 4. Discussion

Generally, this study proposes a new framework to calculate SnoDRI, an index that can measure snow droughts. The framework could be advantageous due to the multiple strengths that we identified. Firstly, it can be applied in ungauged basins. Since we use only the selected features from NLDAS-2 forcings, a gridded dataset integrating satellite observation with measurement gauges and radars, we do not need to rely on any ground-measured variables to calculate SnoDRI. Several basins worldwide are still ungauged, especially in developing countries, leading to a lack of an efficient drought monitoring system. Our framework can act as an alternative snow drought indication framework in such regions. Secondly, our framework reduces the subjectivity in choosing the input variables by using random forest models to select important features. Previous studies examined snow droughts with definitions based on a handful of variables (mostly precipitation, temperature, and SWE) selected based on expert knowledge and assumptions, which add subjectivity to the analysis [1,12]. Thirdly, our framework, besides calculating the index, can give insights into what factors drive snow drought conditions, as represented by the weights from the autoencoder with MI. Typically, studies investigating snow droughts calculate the index based on the abnormal variations in snow variables. Fourthly, regardless of several model inputs that possess multicollinearity, a common issue while using multiple predictor variables, we can see that the SnoDRI framework could eliminate redundant information to a certain extent. For instance, the model gave a lower weight to specific humidity, a variable used along with temperature to calculate snow fraction. Finally, unlike SPI or Palmer Drought Severity Index (PDSI; Palmer, 1965), the proposed ML-based framework is not sensitive to the time series length[14]. Once the model is trained, its weights are fixed and can be applied to a new dataset, no matter the range of time. However, the efficacy of SnoDRI in capturing multi-scale (both spatial and temporal) droughts needs to be explored in greater depth. The abovementioned capabilities of the framework highlight the competence of ML and information theory metrics in assessing snow droughts (or any drought, for that matter).

Formulating the framework for drought index calculation came with several challenges. Most importantly, the absence of a target variable to train and test the ML model. It is not straightforward to establish the performance of the index with a statistical metric (e.g., NSE or KGE). Rather, we had to compare and contrast the indicator variables with the index to see whether it was able to capture the signals of drought. This gave rise to another challenge: the lack of characterization of snow droughts in the present literature. Despite the ongoing interest in the topic, researchers have not reached an established definition of snow drought. This introduces uncertainty in assessing the presence of snow droughts. Related to the above, another challenge in this study was the validation of the new index. It is a common problem in all drought index development studies [46]. Generally, the researchers compare a new index with some of the reported drought events to validate the index [47–49]. Nevertheless, this does not show exclusively the ability of an index to capture all the drought events. Some studies validate their index by checking its congruency with the US Drought Monitor [49,50]. There is vast room for research in developing an established framework for validating drought indices. A possible way to create a validation framework is to verify the closeness of the distributions of relevant variables with that of the index. The closeness of different distributions can be quantified with statistical methods.

We acknowledge that, to some extent, the index is susceptible to the group of input variables we start with before the random forest feature selection. The importance of any variable obtained from the random forest feature selection depends on the whole set of input variables. In other words, adding more variables or choosing a different dataset might produce a different order of feature importance. Although we only considered the NLDAS-2 variables in calculating SnoDRI, the framework can be applied to any set of time series input variables. For example, the ERA5-Land dataset, which provides a larger number of variables with a more extended period [51], and the data from Land Information System (LIS) simulations. In spite of the high computational cost, it would be interesting to see the performance of the framework with ERA-5 or LIS variables. We have executed the framework at a basin scale in the western US. Future efforts can be directed towards establishing the framework in a gridded manner at continental or global scales.

Although we focused on snow droughts, this framework can be applied to identify any type of drought, given the appropriate input variables and feature selection. We selected variables by regressing random forests against SWE and discharge. Training the random forests against different target variables relevant to the interested drought type would give another set of input variables. These can be transformed into an index by following the steps in our framework. Thus, by design, our framework is transferrable. We can set up the framework for any region of interest by training the random forest (for feature selection) and autoencoder (for estimating the weights of selected features through MI scores) with the data of that region. It should be noted that, for different regions, the model might assign different weights to variables depending upon the hydroclimatic characteristics of the region. For instance, in the regions where temperature variability has greater influence, temperature is most likely to control the compressed bottleneck information inside the autoencoder, leading to a higher MI score for temperature (with bottleneck information). This aspect of our framework can be used to get insights about the impact of each variable on droughts.

## 5. Conclusion

We developed a framework to calculate a new index, SnoDRI, that can be used to identify snow droughts. We trained random forest models for 85 basins across the west coast to select the input features for the index calculation. Our novel framework showed the capability of combining autoencoders (a self-supervised machine-learning algorithm) with MI (a degree of mutual dependency between two variables) to estimate the importance of input variables in the occurrence of snow droughts. We found that the downward shortwave radiation, SPIs, temperature, and snow fraction considerably influence snow droughts. In validation, SnoDRI successfully captured the reported snow drought events and their signals in Upper Tuolumne Basin in the Sierra Nevada region. The framework demonstrated the potential to eliminate redundant information in the dataset. The novelty of our framework is that it can be applied to ungauged basins since it does not use any ground measurements. The framework can be adapted to other types of droughts and to different regions around the world.

## Data Availability

The datasets used and/or analyzed in the study are available from the corresponding author on reasonable request.

## Acknowledgments

We would like to thank the Holland Computing Center at the University of Nebraska-Lincoln for providing high-performance computing resources.

## Author Contribution Statement

SRK and TR designed the framework. SRK implemented the framework. SRK, KKK, SS conducted data pre-processing. TR supervised the work. MS and TT provided important feedback. SRK prepared the article with contributions from KKK, SS, TT, MS, and TR.